\documentclass{article}
\usepackage{spconf,amsmath,graphicx}
\usepackage{multirow}
\usepackage{float}
\usepackage{algorithm}
\usepackage{algpseudocode}
\usepackage[fontsize=9pt]{fontsize}
\usepackage{bold-extra}
\usepackage[T1]{fontenc}
\usepackage[subtle]{savetrees}
\usepackage{amssymb}
\usepackage{pifont}
\usepackage{url}
\usepackage{hyperref}
\usepackage{ctable}
\usepackage{cleveref}
\usepackage{booktabs}
\usepackage{tabularx}
\usepackage{makecell}

\newcommand{\comment}[1]{}

\newcommand{\magenta}[1]{{\color{magenta} #1}}

\newcommand{\Tref}[1]{Table~\ref{#1}}

\newcommand{\Fref}[1]{Figure~\ref{#1}}

\newcommand{\app}{\raise.17ex\hbox{$\scriptstyle\sim$}}

\usepackage{amsmath}

\usepackage{colortbl,xcolor}

\usepackage{xspace}
\makeatletter
\def\onedot{\ifx\@let@token.\else.\null\fi\xspace}
\makeatother

\def\etal{\textit{et al}\onedot}

\usepackage{eucal}

\newcommand*{\smfont}{\small}

\title{M{\smfont argin}NCE: Robust Sound Localization with a Negative Margin}

\name{Sooyoung Park$^{1,2}$*, Arda Senocak$^{1}$*, Joon Son Chung$^{1}$ \thanks{*These authors contributed equally to this work.}}
\address{
${}^{1}$Korea Advanced Institute of Science and Technology, South Korea\\
${}^{2}$Electronics and Telecommunications Research Institute, South Korea}

\begin{document}

\maketitle

\begin{abstract}
The goal of this work is to localize sound sources in visual scenes with a self-supervised approach. Contrastive learning in the context of sound source localization leverages the natural correspondence between audio and visual signals where the audio-visual pairs from the same source are assumed as positive, while randomly selected pairs are negatives. However, this approach brings in noisy correspondences; for example, positive audio and visual pair signals that may be unrelated to each other, or negative pairs that may contain semantically similar samples to the positive one. Our key contribution in this work is to show that using a less strict decision boundary in contrastive learning can alleviate the effect of noisy correspondences in sound source localization. We propose a simple yet effective approach by slightly modifying the contrastive loss with a negative margin. Extensive experimental results show that our approach gives on-par or better performance than the state-of-the-art methods. Furthermore, we demonstrate that the introduction of a negative margin to existing methods results in a consistent improvement in performance.
\end{abstract}


\section{Introduction}\label{sec:intro}
We understand the world around us through multiple sensory signals. Among them, sight and sound signals are continuously used for our perception. To make this perception ability seamless, the human brain has developed to organize audio and visual modalities by associating or separating them. Thus, mimicking this ability is of great interest in order to have better learning algorithms. Audio-visual learning is explored with a variety of tasks such as audio-visual fusion and video understanding~\cite{korbar2018cooperative,kazakos2019epicFusion,xiao2020avSlowFast,gao2020listen2look,chung2020seeing,wang2020multimodalHard}, sound source localization~\cite{senocak2018learning,arandjelovic2018objects,senocak2019learning,hu2019deep,qian2020multiple,afouras2020AVObjects,hu2020discriminative,chen2021localizing,senocakHardPos,senocakLessMore,song2022sspl,ezvsl,ssslTransformation}, audio spatialization~\cite{morgadoNIPS18,Yang_2020_CVPR}, and audio-visual sound separation~\cite{afouras2020AVObjects,ephrat2018looking,zhao2019sound,gao2019coSep,gao2021visualVoice,tzinis2021audioScope}.

In this work, we explore the sound source localization task. Human perception has the capability to easily find the objects or events that make the sound in the scene. We leverage the natural correspondence between how objects look and what sounds they make. Humans learn this correspondence without any specific training phase during their daily lives. Thus, accurately solving the sound localization problem in a self-supervised way is the main goal of this line of research. There have been vast efforts on self-supervised sound localization tasks recently.~\cite{senocak2018learning,arandjelovic2018objects,senocak2019learning,Owens2018AudioVisualSA} use the correspondence as a self-supervision proxy task in their training for sound localization. Further,~\cite{senocak2018learning,senocak2019learning} use the attention mechanism to refine visual features with the obtained sound source localization predictions to learn the better correspondence between audio and visual signals. Hu~\etal\cite{hu2019deep} incorporate a clustering approach in audio-visual samples to learn cross-modal correlation. More recently, starting with~\cite{chen2021localizing}, the noise contrastive learning is adopted in sound localization methods~\cite{senocakHardPos,ezvsl,ssslTransformation}. While~\cite{chen2021localizing} uses a hard negative mining approach to consider the background area in the positive image as a negative sample for contrastive learning, Senocak~\etal\cite{senocakHardPos} mine multiple semantically similar samples -- hard positives -- to use in contrastive learning.~\cite{ssslTransformation} extends the model of LVS~\cite{chen2021localizing} with aggressive data augmentations along with the geometrical consistency loss to give invariance and equivariance properties.~\cite{ezvsl} presents a multiple instance learning approach by focusing only on the most correspondent area on the image for audio-visual contrastive learning. Additionally, the initial self-supervised sound localization predictions are refined by mixing the object-guided activation maps from pre-trained visual encoders in the inference phase in~\cite{ezvsl}. Lastly, different from the aforementioned approaches that incorporate positive and negative samples together, ~\cite{song2022sspl} explores a negative-free learning approach in sound localization. 

The correspondence between audio and visual signals undoubtedly plays a key role in general audio-visual learning and specifically in sound source localization as well. Contrastive learning approaches aim to align the features of the same instances while distinguishing the ones from different instances. Similarly, contrastive learning in the context of audio-visual learning leverages the natural correspondence and assigns the audio-visual pairs from the same source as positive and randomly selected mismatched pairs as negative because they are not related. While this seems plausible ideally, it leads to noisy correspondences in reality. These noisy correspondences can be in two forms - (1) Audio-visual signals from the same source may not be semantically related, uninformative to each other, (2) Negative pairs may contain semantically related information to the positive one due to the random selection in a batch. Morgado~\etal\cite{morgado2021robust} show that learning process is falsely guided because of these noisy correspondences when contrastive learning is used without careful consideration.  

This motivates us to design a sound localization method that is more robust to noisy correspondences. To this end, we introduce a ``negative margin'' in the contrastive learning loss function, InfoNCE~\cite{infoNCE}, to reduce the effect of the noisy samples. Considering the standard InfoNCE loss with a zero margin or a positive margin, these noisy correspondences will be pulled or pushed falsely in the wrong direction, even more with a positive margin. It degrades the learning ability of the model. However, as discussed in~\cite{xie2022delving}, a negative margin can alleviate the effect of noisy correspondences by providing a looser decision boundary. Our experiments support our design, employing InfoNCE loss with a negative margin, by showing that this simple approach improves the performance of the sound localization performance on standard benchmarks. 

We propose a new training loss rather than a new sound source localization architecture. To the best of our knowledge, this is the first study on the effect of margin value in contrastive learning loss for sound source localization. Our main contributions are summarized as follows: 1) We present a self-supervised sound source localization model that uses a margin contrastive loss; 2) We demonstrate that using less strict decision boundaries in contrastive learning, only a simple extension of the contrastive loss function with a negative margin, gives on-par or better sound localization performance with state-of-the-art methods that use additional task-oriented strategies; 3) We further investigate that applying a negative margin contrastive loss into existing works consistently improves the performances and shows its merit.
\section{Approach}\label{sec:approach}
\subsection{Preliminaries}\label{sec:preliminaries}
This part describes our contrastive learning method for the audio-visual sound source localization. Let the image frame $\mathbf{v}_{i} \in \mathbb{R}^{3\times H_{v}\times W_{v}}$ and the audio spectrogram $\mathbf{a}_{i} \in \mathbb{R}^{1\times H_{a}\times W_{a}}$ from the $i$-th clip $\mathbf{X}_{i}=\{\mathbf{v}_{i}, \mathbf{a}_{i}\}$. To learn audio-visual correspondence, we use the training method maximizing the similarity between the image representation map $\mathbf{V}_{i} = f_{v} ( \mathbf{v}_{i}; \boldsymbol{\theta}_{v}) \in \mathbb{R}^{c \times h \times w}$ using an image encoder $f_{v}$ and global audio representation $\mathbf{A}_{i} = f_{a} ( \mathbf{a}_{i}; \boldsymbol{\theta}_{a}) \in \mathbb{R}^{c}$ using an audio encoder $f_{a}$. Then, we localize the sound source on the given image with an audio-visual response map $\boldsymbol{\alpha}_{ij} \in \mathbb{R}^{h \times w}$ obtained by using the pixel-wise cosine similarity between the image representation $\mathbf{V}_{i}$ and the globally summarized audio representation $\mathbf{A}_{j}$. We build our model based on recent works~\cite{chen2021localizing, ezvsl}. As a baseline, we use LVS-based loss function~\cite{chen2021localizing} on top of the EZ-VSL~\cite{ezvsl} architecture. The objective function of our baseline is as follows:

\begin{equation}
    S_{i,j} = \frac{1}{\big\lVert \sigma\big(\frac{\boldsymbol{\alpha}_{i,j}-\epsilon}{\beta}\big)\big\rVert_{1}} \Big< \sigma \Big (\frac{\boldsymbol{\alpha}_{i,j}-\epsilon}{\beta} \Big),\enspace \boldsymbol{\alpha}_{i,j}\Big>, 
\end{equation}
\begin{equation}
    \mathcal{L}=-\frac{1}{n}\sum_{i=1}^{n} \enspace \text{log}\frac{e^{S_{i,i}/\tau}}{e^{S_{i,i}/\tau} + \sum_{i\neq j} e^{S_{i,j}/\tau}}, 
    \label{eq:suppressed NCE}
\end{equation}
where $\langle \cdot , \cdot \rangle$ refers Frobenius inner product, $\sigma$ is the sigmoid function for thresholding $\boldsymbol{\alpha}_{i,j}$, $\epsilon$ denotes the thresholding parameter, $\beta$ is the temperature for thresholding, $S_{i,j}$ refers the spatial-wise averaged value of the thresholded audio-visual response map, and $\tau$ is the temperature for contrastive loss. In the inference stage, we can deduce where the sound is visually localized in the paired image using audio-visual response map $\boldsymbol{\alpha}_{ii}$ from the clip $\mathbf{X}_{i}$.
\subsection{Training}\label{sec:training}
Based on the architecture and optimization mentioned above, we introduce marginNCE. In general, margin loss has been applied to make strict decision boundaries among the embedding space by adding a positive margin to the distance between two different embeddings to increase discriminability. However, audio-visual learning may suffer from the faulty positive problem because of the possibility that the image and paired audio are not semantically aligned, and the faulty negative problem due to random sampling in batch configurations~\cite{morgado2021robust}. Therefore, as in~\cite{xie2022delving}, we apply a looser decision boundary by using a negative margin $m$ on (\ref{eq:suppressed NCE}) to alleviate the effect of noisy correspondences from the forementioned two problems. We simply modify the contrastive loss with a margin. The proposed objective function, marginNCE, is as follows:

\begin{equation}
    \mathcal{L}_{marginNCE}=-\frac{1}{n}\sum_{i=1}^{n} \enspace \text{log}\frac{e^{(S_{i,i} - m)/\tau}}{e^{(S_{i,i} - m)/\tau} + \sum_{i\neq j} e^{S_{i,j}/\tau}}. 
    \label{eq:margin NCE}
\end{equation}
\section{Experiments}\label{sec:experiments}
\subsection{Datasets and Evaluation Metrics}\label{sec:dataset}
\noindent\textbf{Datasets.} We train our method on VGGSound~\cite{VGGSound} and SoundNet-Flickr-Training set provided by~\cite{senocak2018learning,senocak2019learning}. VGGSound is an audio-visual dataset containing around ~200K videos. SoundNet-Flickr-training set is the subset of SoundNet-Flickr~\cite{aytar2016soundnet} and it has 144K samples. After training, the sound localization performance is tested with VGG-SS~\cite{chen2021localizing} and SoundNet-Flickr-Test~\cite{senocak2018learning} datasets. These evaluation sets have bounding box annotations of sound sources for \app 5K and 250 samples, respectively.

\noindent\textbf{Evaluation metrics.} We measure the sound localization performance with two commonly used metrics: 1) Consensus Intersection over Union (cIoU)~\cite{senocak2018learning} measures the localization accuracy between the ground-truth and the prediction with intersection over union approach. 2) Area Under Curve (AUC) measures the area under the cIoU curve plotted by various threshold values from 0 to 1. 

\vspace{-4mm}
\subsection{Implementation Details}\label{sec:implementation}
Following the common practice in earlier sound localization methods~\cite{chen2021localizing,senocakHardPos,ezvsl,ssslTransformation}, we use the center frame of the video with the corresponding 3 seconds audio segment around that frame as input data during training on the VGGSound dataset. In the SoundNet-Flickr dataset, frames are given with the paired audio. Input images for training are in the size of $224 \times 224$. We use 16kHz sampling rate for audios in both datasets. We transform audios to log spectrograms with the size of $257 \times 200$. Similar to~\cite{chen2021localizing,senocakHardPos,ezvsl,ssslTransformation}, ResNet18 is used as a backbone network for each modality. We set the hyperparameters as $\epsilon =0.65$, $\beta = 0.03$, and $\tau= 0.07$. Unless it is mentioned explicitly, we adopt the value of -0.2 as a margin in our experiments. We use adam optimizer with the weight decay. We train our model for 20 epochs.
\subsection{Quantitative Results}\label{sec:quantitative}

\begin{table}[t!]
\centering
\resizebox{0.8\linewidth}{!}{
\begin{tabular}{lcccc}
\toprule
& \multicolumn{2}{c}{\textbf{VGG-SS}} & \multicolumn{2}{c}{\textbf{Flickr-SoundNet}} \\
\textbf{Method} & \textbf{cIoU $\uparrow$} & \textbf{AUC $\uparrow$} & \textbf{cIoU $\uparrow$} & \textbf{AUC $\uparrow$} \\ \midrule
Attention~\cite{senocak2018learning}$_{\text{CVPR}18}$  &  18.50 & 30.20 & 66.00 & 55.80 \\
LCBM~\cite{senocakLessMore}$_{\text{WACV}22}$  	 & 32.20 & 36.60 & - & - \\
LVS~\cite{chen2021localizing}$\dagger$$_{\text{CVPR}21}$   & 30.30 & 36.40 & 72.40 & 57.80 \\
LVS~\cite{chen2021localizing}$_{\text{CVPR}21}$   & 34.40 & 38.20 & 71.90 & 58.20 \\
HardPos~\cite{senocakHardPos}$_{\text{ICASSP}22}$   & 34.60 & 38.00 & 76.80 & 59.20 \\
SSPL(w/o PCM)~\cite{song2022sspl}$_{\text{CVPR}22}$   & 27.00 & 34.80 & 73.90 & 60.20 \\
SSPL(w/ PCM)~\cite{song2022sspl}$_{\text{CVPR}22}$   & 33.90 & 38.00 & 76.70 & 60.50 \\
EZ-VSL(w/o OGL)~\cite{ezvsl}$_{\text{ECCV}22}$   & 35.96 & 38.20 & 78.31 & 61.74 \\
SSL-TIE~\cite{ssslTransformation}$_{\text{ACM MM}22}$   & \textbf{38.63} & \textbf{39.65} & 79.50 & 61.20 \\
\rowcolor{lightgray!25}
\textbf{Ours}& 38.25 & 39.06 & \textbf{83.94} & \textbf{63.20} \\
\bottomrule
EZ-VSL(w/ OGL.)~\cite{ezvsl}$_{\text{ECCV}22}$   & 38.85 & 39.54 & 83.94 & 63.60 \\
\rowcolor{lightgray!25}
\textbf{Ours (w/ OGL)}& \textbf{39.78} & \textbf{40.01} & \textbf{85.14} & \textbf{64.55} \\
\bottomrule
\end{tabular}}

{
\caption{\textbf{Quantitative results on the VGG-SS and SoundNet-Flickr test sets}. All models are trained with 144K samples from VGG-Sound and tested on VGG-SS and SoundNet-Flickr. $\dagger$ is the result of the model released on the official project page.}\label{tab:quantitative}}
\vspace{-2mm}
\end{table}

\begin{table}[tb!]
\centering
\resizebox{0.7\linewidth}{!}{
\begin{tabular}{lcc}
\toprule
\textbf{Method}        		    & \textbf{cIoU $\uparrow$}    & \textbf{AUC $\uparrow$} \\ \midrule
Attention\cite{senocak2018learning}$_{\text{CVPR}18}$ & 66.00 & 55.80 \\
DMC\cite{hu2019deep}$_{\text{CVPR}19}$ & 67.10 & 56.80 \\
LVS~\cite{chen2021localizing}$\dagger$$_{\text{CVPR}21}$  	       & 67.20 & 56.20 \\
LVS~\cite{chen2021localizing}$_{\text{CVPR}21}$  	       & 69.90 & 57.30 \\
HardPos~\cite{senocakHardPos}$_{\text{ICASSP}22}$ & 75.20 & 59.70 \\
SSPL(w/o PCM)~\cite{song2022sspl}$_{\text{CVPR}22}$   & 69.90 & 58.00 \\
SSPL(w/ PCM)~\cite{song2022sspl}$_{\text{CVPR}22}$   & 75.90 & 61.00 \\
EZ-VSL(w/o OGL)~\cite{ezvsl}$_{\text{ECCV}22}$   & 71.89 & 58.81 \\
SSL-TIE~\cite{ssslTransformation}$_{\text{ACM MM}22}$   & 81.50 & 61.10 \\
\rowcolor{lightgray!25}
\textbf{Ours}& \textbf{84.74} & \textbf{63.08} \\
\bottomrule
EZ-VSL(w/ OGL)~\cite{ezvsl}$_{\text{ECCV}22}$   & 83.13 & 63.06 \\
\rowcolor{lightgray!25}
\textbf{Ours(w/ OGL)}& \textbf{85.54} & \textbf{64.27} \\
\bottomrule
\end{tabular}}
{
\caption{\textbf{Quantitative results on the SoundNet-Flickr test set.} All models are trained and tested on the SoundNet-Flickr dataset. $\dagger$  is the result of the model from the official project page.}\label{tab:quantitative_second}}
\vspace{-2mm}
\end{table}

\subsubsection{Comparison with the State-of-the-art Methods}
In this section, we compare our method with existing sound source localization approaches. Specifically, we provide the results in two settings by following previous works~\cite{chen2021localizing,ezvsl,ssslTransformation}: 1) Training on VGGSound-144K and testing on VGG-SS and SoundNet-Flickr test sets 2) Training on SoundNet-Flickr-Training-144K and testing on SoundNet-Flickr test set. All the models are trained and tested on the same amount of data. Results are shown in~\Tref{tab:quantitative} and~\Tref{tab:quantitative_second}. Our model outperforms prior work on the SoundNet-Flickr test set regardless of datasets it is trained, 4.44$\%$ cIoU and 1.46$\%$ AUC when trained on VGGSound and 3.24$\%$ cIoU and 1.98$\%$ AUC when trained on SoundNet-Flickr. However, it achieves slightly lower accuracy compared to~\cite{ssslTransformation} on the VGG-SS test when trained on VGGSound. We would like to highlight that existing approaches use additional task-specific strategies such as; SSL-TIE~\cite{ssslTransformation} incorporates aggressive augmentations and transformation together with additional geometrical consistency loss and background suppression, SSPL~\cite{song2022sspl} attaches an explicit sub-module called PCM to reduce the effect of background noise, and EZ-VSL~\cite{ezvsl} refines their initial localization results by using object guidance (OGL) in the inference stage. In contrast, our model only uses a simple approach that extends the training objective with a negative margin and it still gives an on-par or better performance with the existing state-of-the-art methods. As aforementioned, EZ-VSL proposes a refinement of audio-visual sound localization with object-guided localization (OGL). To make a fair comparison with EZ-VSL, we also report the performance of our method with OGL, and the results are shown in the bottom part of the tables. Note that our method gives an on-par performance with EZ-VSL (OGL) even \emph{without using OGL}. We do not use OGL in our architecture in the remainder of this paper unless it is directly compared with EZ-VSL (OGL). 

\subsubsection{Cross-Dataset Audio-Visual Localization}
As expected, the best results are typically obtained when training and testing are done on the same dataset. Here, we present the cross-dataset generalization performance where the datasets used for training and testing are different.~\Tref{tab: exp_cross} shows the quantitative results where the model is trained on VGGSound-144K and SoundNet-Flickr-144K, and tested on SoundNet-Flickr and VGG-SS test sets respectively. As the results show, our model has better generalization ability and it outperforms all the other methods in this task.  

\begin{table}[!tb]
	\renewcommand\tabcolsep{6.0pt}
	\centering
	\scalebox{0.68}{
		\begin{tabular}{cclcc}
			\toprule
			Test Set & Training Set & Method & cIoU $\uparrow$ & AUC $\uparrow$  \\ 	
			\midrule
             \multirow{5}{*}{SoundNet-Flickr} & \multirow{5}{*}{VGGSound 144k} & LVS~\cite{chen2021localizing}$_{\text{CVPR}21}$ & 71.90 & 58.20 \\
             & & EZ-VSL(w/o OGL)~\cite{ezvsl}$_{\text{ECCV}22}$ & 78.31 & 61.74 \\
             & & \textbf{Ours} & \textbf{83.94} & \textbf{63.20} \\
             \cline{3-5}
             & & EZ-VSL(w/ OGL)~\cite{ezvsl}$_{\text{ECCV}22}$ & 83.94 & 63.60 \\
             & & \textbf{Ours(w/ OGL)} & \textbf{85.14} & \textbf{64.55} \\ \hline
             \multirow{5}{*}{VGG-SS} & \multirow{5}{*}{SoundNet-Flickr 144k} & LVS~\cite{chen2021localizing}$_{\text{CVPR}21}$ & 26.95 & 34.30 \\
             & & EZ-VSL(w/o OGL)~\cite{ezvsl}$_{\text{ECCV}22}$ & 29.39 & 35.53 \\
             & & \textbf{Ours} & \textbf{34.45} & \textbf{37.35} \\
             \cline{3-5}
             & & EZ-VSL(w/ OGL)~\cite{ezvsl}$_{\text{ECCV}22}$ & 38.62 & 39.20 \\
             & & \textbf{Ours(w/ OGL)} & \textbf{39.41} & \textbf{39.81} \\
             \bottomrule
			\end{tabular}}
\caption{\textbf{Quantitative results for cross-dataset evaluation.}}\label{tab: exp_cross}
\vspace{-2mm}
\end{table}

\vspace{-2mm}
\subsubsection{Open-Set Audio-Visual Localization} 
Another assessment we can conduct on the generalization ability of our model is to evaluate the model in open-set settings where testing samples come from categories that are not used during self-supervised training. Here, following the train/test splits of previous works~\cite{chen2021localizing,ezvsl}, the model is trained with randomly selected 110 categories from VGGSound. Then, the evaluation is done on two test sets: 1) \textbf{Heard} shares the same categories with the training set, and 2) \textbf{Unheard} contains 110 disjoint categories from the training set. These categories are never seen and heard by the model during training.

Results are shown in~\Tref{tab: exp_openset_first}. Three phenomena can be observed from this table. First, our method, regardless of the value used for margin, outperforms compared methods in both heard and unheard setups. Second, similar to EZ-VSL, ours also performs better on unheard categories than heard ones. This shows the generalization ability of our method. Third, we see that the negative margin works best among the other margins used in the unheard scenario, which is a real open-set setup. This observation is similar to the findings of~\cite{liu2020negative} that a negative margin is more proper than a positive or zero margin for open-set scenarios. We can also see that while positive margin performance is higher than a zero margin in the heard scenario, it is the opposite in unheard setup which positive margin hurts the discriminability of unseen categories as discussed in~\cite{liu2020negative}. 

\begin{table}[tbp]
\centering
\footnotesize
\scalebox{0.96}{
\begin{tabular}{l|lccc}
\toprule
\textbf{Test Class} & \textbf{Method}   & \textbf{Margin}  & \textbf{cIoU $\uparrow$}  & \textbf{AUC $\uparrow$} \\ 
\hline 
\multirow{6}{*}{Heard 110}    
& LVS~\cite{chen2021localizing}$_{\text{CVPR}21}$ & - &28.90 & 36.20 \\
& EZ-VSL(w/o OGL)~\cite{ezvsl}$_{\text{ECCV}22}$ & - &31.86 & 36.19 \\
\cline{2-5}
& \textbf{Ours} & -0.2 &\textbf{36.35} & \textbf{37.92}\\
& Ours & 0.0 &34.40 & 37.38\\
& Ours & +0.2 &34.83 & 37.50\\
\cline{2-5}
& EZ-VSL(w/ OGL)~\cite{ezvsl}$_{\text{ECCV}22}$ & - &37.25 & 38.97\\ 
& \textbf{Ours(w/ OGL)} & -0.2 &\textbf{38.07} & \textbf{39.39} \\
\hline \hline
\multirow{7}{*}{Unheard 110} 
& LVS~\cite{chen2021localizing}$_{\text{CVPR}21}$ & - &26.30 & 34.70 \\
& EZ-VSL(w/o OGL)~\cite{ezvsl}$_{\text{ECCV}22}$ & - &32.66 & 36.72 \\
\cline{2-5}
& \textbf{Ours} & -0.2 &\textbf{37.90} & \textbf{39.17}\\
& Ours & 0.0 &36.74 & 38.39\\
& Ours & +0.2 &36.15 & 38.44\\
\cline{2-5}
& EZ-VSL(w/ OGL)~\cite{ezvsl}$_{\text{ECCV}22}$ & - &39.57 & 39.60 \\
& \textbf{Ours(w/ OGL)} & -0.2 &\textbf{40.58} & \textbf{40.30}\\ 
\bottomrule
\end{tabular}}
\caption{\textbf{Comparison results on open-set audio-visual localization experiments trained and tested on the splits of~\cite{chen2021localizing,ezvsl}.}} \label{tab: exp_openset_first}
\end{table}

\begin{table}[tb]
\centering
\resizebox{0.8\linewidth}{!}{
\begin{tabular}{lccc}
\toprule
\textbf{Method} & \textbf{Margin}        		    & \textbf{cIoU $\uparrow$}    & \textbf{AUC $\uparrow$} \\ \midrule
LVS~\cite{chen2021localizing}$_{\text{CVPR}21}$  & 0.0	       & 33.99 & 37.76 \\
\rowcolor{lightgray!25}
LVS-marginNCE  & -0.2	       & \textbf{34.80} & \textbf{38.17} \\
\rowcolor{lightgray!25}
\textbf{LVS-marginNCE}  & -0.3	       & \textbf{35.73} & \textbf{38.52} \\
\bottomrule
EZ-VSL(w/o OGL)~\cite{ezvsl}$_{\text{ECCV}22}$ & 0.0   & 37.57 & 38.70 \\
\rowcolor{lightgray!25}
\textbf{EZ-VSL(w/o OGL)-marginNCE} & -0.2  & \textbf{38.70} & \textbf{39.26} \\
\bottomrule
Ours & 0.0 & 36.93 & 38.58 \\
\rowcolor{lightgray!25}
\textbf{Ours-marginNCE} & -0.2 & \textbf{38.25} & \textbf{39.06} \\
\bottomrule
\end{tabular}
}

\caption{\textbf{Generalization of the marginNCE on different baselines.} All models are trained with 144K samples from VGG-Sound and tested on VGG-SS.}\label{tab:quantitative_marginNCE_benchmark}
\vspace{-2mm}
\end{table}

\begin{table*}[ht!]
\footnotesize
\scalebox{0.9}{
\begin{tabular}{l c cc c cc c cc c cc c cc c cc}
\Xhline{2\arrayrulewidth}
\multirow{2.5}{*}{\textbf{Dataset}} & \phantom{a} & \multicolumn{2}{c}{\textbf{{\scriptsize margin 0.2}}} & \phantom{a} & \multicolumn{2}{c}{\textbf{{\scriptsize margin 0.0}}} & \phantom{a} & \multicolumn{2}{c}{\textbf{{\scriptsize margin -0.1}}} & \phantom{a} & \multicolumn{2}{c}{\textbf{{\scriptsize margin -0.2}}} & \phantom{a} & \multicolumn{2}{c}{\textbf{{\scriptsize margin -0.3}}} & \phantom{a} & \multicolumn{2}{c}{\textbf{{\scriptsize margin -0.4}}} \\
\cmidrule{3-4} \cmidrule{6-7} \cmidrule{9-10} \cmidrule{12-13} \cmidrule{15-16} \cmidrule{18-19}
 && \textbf{cIoU $\uparrow$} & \textbf{AUC $\uparrow$} && \textbf{cIoU $\uparrow$} & \textbf{AUC $\uparrow$} && \textbf{cIoU $\uparrow$} & \textbf{AUC $\uparrow$} && \textbf{cIoU $\uparrow$} & \textbf{AUC $\uparrow$} && \textbf{cIoU $\uparrow$} & \textbf{AUC $\uparrow$} && \textbf{cIoU $\uparrow$} & \textbf{AUC $\uparrow$}\\
\hline 
VGG-SS &&  37.05 & 38.56 && 36.93 & 38.58 &&  37.51 & 38.89 && 38.25 & 39.06 &&  37.65 & 38.84 && 36.93 & 38.47   \\
\hline 

SoundNet-Flickr &&  82.73 & 62.44 && 82.73 & 62.64 &&  81.93 & 62.46 && 83.94 & 63.20 &&  83.53 & 63.70 && 84.74 & 63.00 \\
\Xhline{2\arrayrulewidth}
\end{tabular}} 
\caption{\textbf{Accuracy w.r.t. Different Margins.} The results show that performance is improved by setting an appropriate negative margin.}\label{tab:ablation}
\normalsize
\vspace{-2mm}
\end{table*}
\begin{figure*}[htb!]
\centering
{
\resizebox{1\linewidth}{!}{%
\begin{tabular}{c}
\includegraphics[width = 1\linewidth]{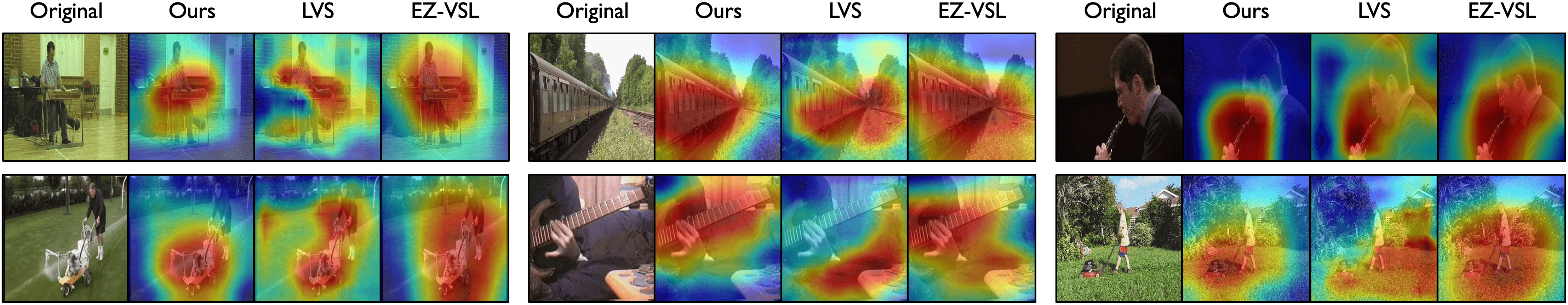} \\
\end{tabular}
}
}
\caption{\textbf{Sound localization results on VGG-SS and comparison with the state-of-the-art methods~\cite{chen2021localizing,ezvsl}.}}
\label{fig:qualitative_vggss}
\vspace{-2mm}
\end{figure*}

\begin{figure*}[htb!]
\centering
{
\resizebox{1\linewidth}{!}{%
\begin{tabular}{c}
\includegraphics[width = 1\linewidth]{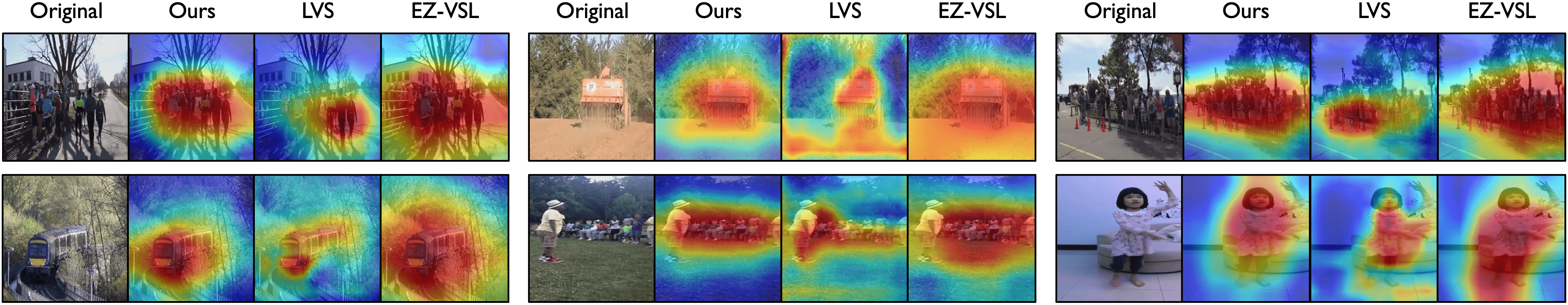} \\
\end{tabular}
}
}
\caption{\textbf{Sound localization results on SoundNet-Flickr and comparison with the state-of-the-art methods~\cite{chen2021localizing,ezvsl}.}}
\label{fig:qualitative_flickr}
\vspace{-2mm}
\end{figure*}

\subsubsection{Generalization of the Negative Margin on Different Baselines}
To demonstrate that negative marginNCE is generally applicable to the other sound localization methods, we conduct experiments with the most recent methods by extending their loss with a negative margin. All of these baselines use InfoNCE loss in their architectures. We train each baseline with their publicly released codes on VGGSound-144K and test on the VGG-SS dataset for a fair comparison. Results in~\Tref{tab:quantitative_marginNCE_benchmark} show that using a negative margin consistently improves upon all baselines. It even helps EZ-VSL to get the state-of-the-art performance on this experimental setup (compare the results with~\Tref{tab:quantitative}).

\subsubsection{Comparison of the Different Decision Margins}
We conduct an ablation study to explore the accuracy of our method w.r.t. different margin values. We show the performance of our model in~\Tref{tab:ablation} when it is trained on VGGSound-144K and tested on VGG-SS and SoundNet-Flickr with different margins. As we expect, we get higher cIoU accuracy when the margin is set to negative than zero or positive margins on both test sets. Similarly, AUC performance is also higher with negative margins on both datasets. 

\subsection{Qualitative Results}\label{sec:qualitative}
In this section, we visualize our sound localization results on VGG-SS and SoundNet-Flickr and compare them with other existing methods~\cite{chen2021localizing,ezvsl}. More results, including failure cases, are available at \href{https://sites.google.com/view/marginssl}{\magenta{https://sites.google.com/view/marginssl}}.

\noindent\textbf{VGG-SS}: We provide the sound localization results of VGG-SS samples in~\Fref{fig:qualitative_vggss}. Our results are more accurate and compact than the other methods. Our qualitative results show that our method handles the co-occurring class-related backgrounds or objects better than the other methods. Green grass background/land or humans often co-occur in the images of ``lawn mowers''. While our method localizes the lawn mower accurately, the response maps of the other methods contain human and green grass areas as well.  As seen in the ``man plays a flute'' example (first row and third column), our method only focuses on the location of the flute, not on the man. However, EZ-VSL contains the area where the human head exists. A similar trend can be also seen in the example of ``man plays a slide guitar'' (first row and first column).

\noindent\textbf{SoundNet-Flickr}: Our results in~\Fref{fig:qualitative_flickr} depict more accurate localization responses in comparison to the recent methods in the SoundNet-Flickr test set as well. We notice that our method gives more accurate results for the scenes with the ``crowd''. While LVS results can not cover the entire crowd, EZ-VSL results generally contain a larger area than the crowd itself.
\section{Conclusion}\label{sec:conclusion}
In this paper, we concentrate on the problem of self-supervised
sound source localization with contrastive learning. We identify noisy correspondences due to the assumption of a natural correspondence between audio and visual signals in contrastive learning. With the motivation that looser decision boundaries can alleviate the effect of these noisy correspondences on training, we suggest a simple extension of the contrastive loss function with a negative margin without bells and whistles. Our experiments support our design by showing on-par or state-of-the-art performance on standard benchmarks. We further demonstrate that the proposed negative margin is applicable to any existing approach with contrastive loss and their performances are consistently improved. Therefore, audio-visual sound source localization studies can benefit from our work.

\section{Acknowledgments}\label{sec:acknowledgements}
This work was supported by Electronics and Telecommunications Research Institute (ETRI) grant funded by the Korean government, [22ZH1200, The research of the basic media$\cdot$contents technologies].

\newpage
\bibliographystyle{IEEEbib}
\bibliography{shortstrings,refs}

\end{document}